\documentclass[runningheads]{llncs}
\usepackage{amsmath}
\usepackage{amssymb}
\usepackage{float}
\usepackage{amsfonts}
\usepackage{subcaption}
\usepackage{wrapfig}
\usepackage[all]{nowidow}
\usepackage[table,xcdraw]{xcolor}

\DeclareMathOperator*{\argmin}{arg\,min}
\usepackage[T1]{fontenc}
\usepackage{graphicx}
\usepackage{multirow}
\usepackage[colorlinks=true, linkcolor=blue, citecolor=blue, urlcolor=blue]{hyperref}
\usepackage{color}

\usepackage{cleveref}

\newcommand\blfootnote[1]{%
  \begingroup
  \renewcommand\thefootnote{}\footnote{#1}%
  \addtocounter{footnote}{-1}%
  \endgroup
}

\begin{document}
\title{Fast Cross-Strength Multi-Contrast Brain MRI Translation using Latent Bridge Matching}
\titlerunning{Fast Cross-Strength MRI Translation}
%
\author{Siddharth Srivastava \and Till Bretschneider}
%
\authorrunning{S. Srivastava \and T. Bretschneider}
%
\institute{University of Warwick, United Kingdom \\\email{\{Siddharth.Srivastava,Till.Bretschneider\}@warwick.ac.uk}}
%
\maketitle              
\begin{abstract}
    Magnetic Resonance Imaging (MRI) acquired at different field strengths exhibits pronounced variation in noise, resolution, homogeneity, and contrast, which limits comparability across acquisition settings and complicates downstream analysis. We address this with a unified conditional model for controllable field-to-field synthesis, built on the framework of conditional latent bridge matching. Our single model achieves highly competitive results across the validation phase for all three tasks of the MRIxFields2026 challenge without task-specific architectures or training. We achieve fast generation with only a single inference step, producing all modality and field-strength combinations for $30$ axial slices in under $90$ seconds, as well as cross-modality-strength translation for a full volume in under $70$ seconds, on a single NVIDIA A5000 GPU. We further provide extensive ablations regarding different components of our solution. \blfootnote{Code: \href{https://gitlab.com/siddharthsrivastava/mrixfields-2026}{https://gitlab.com/siddharthsrivastava/mrixfields-2026}}
\end{abstract}

\section{Introduction}
Magnetic Resonance Imaging (MRI) is a widely used method for non-invasive medical 3D imaging, providing both tissue and anatomical contrast for clinical diagnosis and medical research. MRI quality is highly dependent on the magnetic field strength of the acquisition scanner, with strengths ranging from ultra-low-field portable scanners at 0.1T, to clinical scanners in point-of-care settings at 1.5--3T, to ultra-high-field research scanners at 7T. Images acquired at different strengths vary substantially regarding geometric artefacts, signal-to-noise, resolution, homogeneity, and contrast, significantly affecting multi-site studies and the clinical translation of downstream machine learning models trained on such varied data. 

Generative machine learning models based on flow or diffusion frameworks have the potential to infer high-field-equivalent anatomical structures from low-field inputs, overcoming physical hardware gaps. The Generalizable Cross-Field MRI Translation and Harmonization Challenge (MRIxFields2026) aims to provide a single benchmark for cross-field harmonisation and synthesis and advance multimodal clinical translation. The dataset consists of MRI volumes from 0.1T to 7T using T1, T2, and T2FLAIR modalities, split into two partitions: a large unpaired set of 1,900+ volumes, and a smaller paired set of 40 travelling-cohort volunteers (3 for training, 17 for validation, and 20 for testing); each volunteer in the paired set is scanned across all modalities and field strengths. 

We propose a single unified conditional framework for all three tasks of the MRIxFields2026 challenge: ultra-high-field (7T) synthesis from arbitrary input field strengths (0.1T, 1.5T, 3T, 5T), higher-field (1.5T, 3T, 5T, 7T) generation from ultra-low-field MRI, and controllable field-to-field MRI synthesis. Our method encodes the source, target, and auxiliary modality slices into a shared latent space using a variational encoder; a separate lightweight condition encoder encodes the source field strength, target field strength, and target modality. A drift model then learns the conditional latent flow to transport the source latent toward the target latent; the resulting target latent is decoded back to pixel space using a variational decoder. At inference time, our method predicts the target latent in a single ODE step, enabling fast image synthesis: in practice, we generate all $4 \times 3$ field and modality combinations for $30$ axial slices of a single volume in approximately 90 seconds on a single NVIDIA A5000 GPU.

\section{Background}
\subsection{Rectified Flow and Bridge Matching}
Rectified flow \cite{liu2022flowstraightfastlearning} and bridge matching \cite{chadebecLBMLatentBridge2025} are methods that transport a source distribution onto a target distribution by constructing deterministic (stochastic in the bridge matching case) interpolants between samples and estimating the drift of the ODE (SDE in the bridge matching case) of the process using a neural network. Consider two distributions $\pi_0$ and $\pi_1$ from which we sample $X_0 \sim \pi_0$, $X_1 \sim \pi_1$ independently. Rectified flow constructs the linear interpolant 
\[ X_t = (1-t)X_0 + tX_1, \qquad t \in [0,1] \]
which recovers $X_0$ at $t = 0$ and $X_1$ at $t = 1$. The time derivative of $X$ is constant along the interpolant 
\begin{equation}
    \dot{X}_t = \frac{\mathrm{d}}{\mathrm{d}t}[X_t] = X_1 - X_0,
    \label{eqn:flow_ode}
\end{equation}
and describes the velocity to follow a straight path \cite{liu2022flowstraightfastlearning}. This velocity field is modelled as a neural network $v_{\theta}(X, t)$ using a least-squares objective: 
\begin{equation}
    \theta^{\star} := \argmin_{\theta} \mathbb{E}_{t, X_0, X_1} \left[\left\lVert v_{\theta}(X, t) - (X_1 - X_0)\right\rVert^2 \right], \qquad t \sim \tau(t),
\end{equation}
where $\tau(t)$ is the timestep distribution -- whose minimiser is the conditional mean of the line directions $X_1 - X_0$ that pass through $x$ at time $t$
\[ v_{\theta^{\star}}(x,t) \approx \mathbb{E}[X_1 - X_0 \mid X_t = x]. \]
Bridge matching \cite{chadebecLBMLatentBridge2025} instead models the stochastic interpolant $X_t$ as a general Brownian bridge between endpoints $X_0$ and $X_1$: 
\begin{equation}
    X_t = (1-t)X_0 + tX_1 + \sigma B_t, \qquad t \in [0,1]
    \label{eqn:br_bridge}
\end{equation}
where $B_t$ is a standard Brownian bridge process and $\sigma \geq 0$ is the diffusion coefficient; note that $B_t = \sqrt{t(1-t)} \epsilon, \epsilon \sim \mathcal{N}(0, I)$ using the reparameterisation trick, and is pinned at both ends $B_0 = B_1 = 0$. In this case, the dynamics are governed by a stochastic differential equation (SDE), rather than an ordinary differential equation (ODE) as in \cref{eqn:flow_ode}:
\begin{equation}
    \mathrm{d}X_t = \frac{(X_1 - X_t)}{1 - t}\,\mathrm{d}t + \sigma \mathrm{d}W_t
    \label{eqn:bm_sde}
\end{equation}
where $W_t$ is a Brownian motion. As before, we fit a neural network $v_{\theta}(X, t)$ (the \emph{drift} network) to model the drift term using least squares,
\begin{equation}
    \argmin_{\theta} \mathbb{E}_{t, X_0, X_1} \left[\left\lVert v_{\theta}(X_t, t) - \frac{(X_1 - X_t)}{1 - t}\right\rVert^2 \right]
    \label{eqn:bm_loss}
\end{equation}
which recovers the marginal drift $\mathbb{E}\!\left[\frac{X_1 - X_t}{1-t} \middle| X_t\right]$ independent of $X_1$. 

\subsection{Inference}
\label{sec:inference}
After training $v_{\theta}$ using \cref{eqn:bm_loss}, it approximates the marginal $\mathbb{E}\!\left[\frac{X_1 - X_t}{1-t} \,\middle|\, X_t\right]$, which no longer depends on the unknown endpoint $X_1$ as in \cref{eqn:bm_sde}. Therefore, we can generate a sample from $\pi_1$ by first sampling $X_0 \sim \pi_0$ and integrating the learnt SDE
\[ \mathrm{d}X_t = v_\theta(X_t, t)\,\mathrm{d}t + \sigma\, \mathrm{d}W_t\]
from $t=0$ to $t=1$, where $W_t$ is a standard Brownian motion. This integral can be discretised and solved using Euler--Maruyama: on an interval
$0 = \tau_0 < \tau_1 < \dots < \tau_N = 1$ with step size $\Delta t = \tau_{n+1} - \tau_n$, we initialise $\hat{X}_0 \sim \pi_0$ and iterate
\begin{equation}
    \hat{X}_{n+1} = \hat{X}_n + v_\theta(\hat{X}_n, \tau_n)\,\Delta t + \sigma\,\Delta W_n,
    \qquad \Delta W_n \sim \mathcal{N}(0, \Delta t),
    \label{eqn:em_update}
\end{equation}
where the increments $\Delta W_n$ are those of a Brownian motion rather than a bridge, since the endpoint is no longer fixed at inference. The final iterate $\hat{X}_1$ approximates a sample drawn from $\pi_1$. 


\section{Method}
Performing bridge matching directly in pixel space is computationally expensive at high resolution. We therefore adapt the latent-diffusion paradigm \cite{rombach2022highresolutionimagesynthesislatent} and perform bridge matching in a smaller learnt latent space, called \emph{latent} bridge matching \cite{chadebecLBMLatentBridge2025}. Specifically, an encoder maps an image to a lower-dimensional latent space, the generative process is run in latent space, and a decoder reconstructs the image in pixel space. 

In our approach we finetune a pretrained variational autoencoder (VAE) \cite{kingma2022autoencodingvariationalbayes}, consisting of an encoder $\mathcal{E}$ and a decoder $\mathcal{D}$, on 2D axial slices from the large unpaired dataset. Formally, the samples $X_0 \sim \pi_0$ and $X_1 \sim \pi_1$ are encoded to a latent space $Z_0 := \mathcal{E}(X_0), Z_1 := \mathcal{E}(X_1)$. Then, we adapt \cref{eqn:br_bridge} to form the \emph{latent} stochastic interpolant $Z_t$
\begin{equation}
    Z_t = (1-t)Z_0 + t(Z_1) + \sigma B_t, \qquad t \in [0,1]
    \label{eqn:latent_interpol}
\end{equation}
which then leads to the latent bridge matching (LBM) loss based on \cref{eqn:bm_loss}
\begin{equation}
    \mathcal{L}_{\text{LBM}} = \mathbb{E}_{t \sim \tau(t), X_0, X_1} \left[\left\lVert v_{\theta}(Z_t, t) - \frac{\left(Z_1 - Z_t\right)}{1 - t}\right\rVert^2 \right].
    \label{eqn:lbm_loss}
\end{equation}
Inference in latent space follows \cref{sec:inference}, except that we use the dynamics of $\mathrm{d}Z_t = v_\theta(Z_t, t)\,\mathrm{d}t + \sigma\,\mathrm{d}W_t$, and the final approximation $\hat{Z_1}$ is decoded out of latent space as $\hat{X}_1 := \mathcal{D}(\hat{Z}_1)$. 

\subsection{Models}
\begin{figure}[t]
    \centering
    \includegraphics[width=1.0\linewidth]{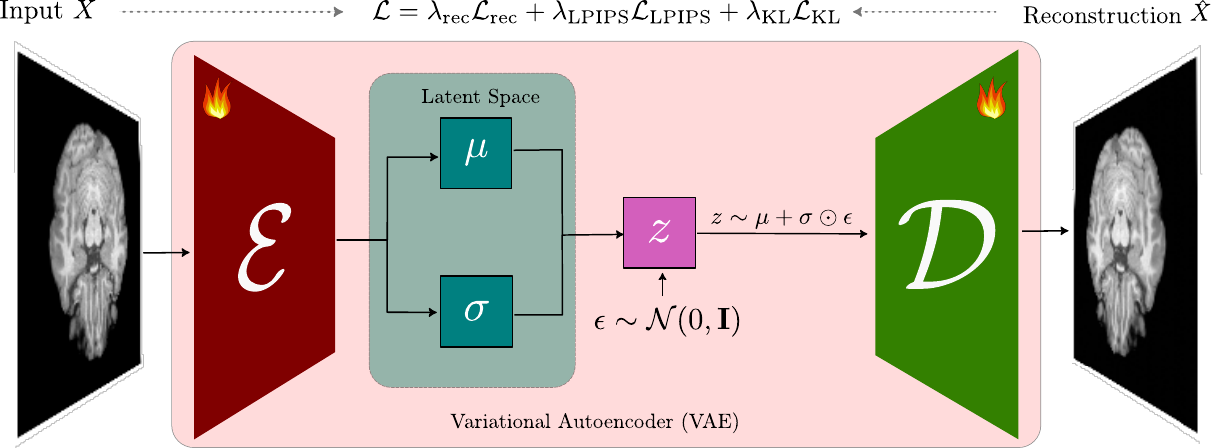}
    \caption{Training pipeline for the variational autoencoder.}
    \label{fig:vae}
\end{figure}

\paragraph{Variational Autoencoder.} 
We initialise a variational autoencoder from the \texttt{sd-\\research/stable-diffusion-2-1-base} checkpoint and finetune on the unpaired dataset: we showcase this training in \Cref{fig:vae}. Given an input slice $X$, the encoder predicts the parameters of the approximate Gaussian posterior parameterised by $\mu$ and $\sigma$, from which the latent code $z$ is sampled using the reparameterisation trick \cite{kingma2022autoencodingvariationalbayes} and passed to the decoder for reconstructing $\hat{X}$. The VAE is optimised end-to-end using the objective 
\begin{equation}
    \mathcal{L} = \lambda_{\text{rec}} \mathcal{L}_{\text{rec}} + \lambda_{\text{LPIPS}} \mathcal{L}_{\text{LPIPS}} + \lambda_{\text{KL}} \mathcal{L}_{\text{KL}}
    \label{eqn:full_vae_loss}
\end{equation}
where $\mathcal{L}_{\text{rec}}$ is the pixel-wise reconstruction loss, $\mathcal{L}_{\text{LPIPS}}$ is the learned perceptual image patch similarity (LPIPS) \cite{zhang2018perceptual} loss, and $\mathcal{L}_{\text{KL}}$ is the reverse KL divergence \cite{KullbackLeibler1951} between $\mathcal{N}(\mu, \sigma^2)$ and $\mathcal{N}(0, \mathbf{I})$.

\begin{figure}[t]
    \centering
    \includegraphics[width=1.0\linewidth]{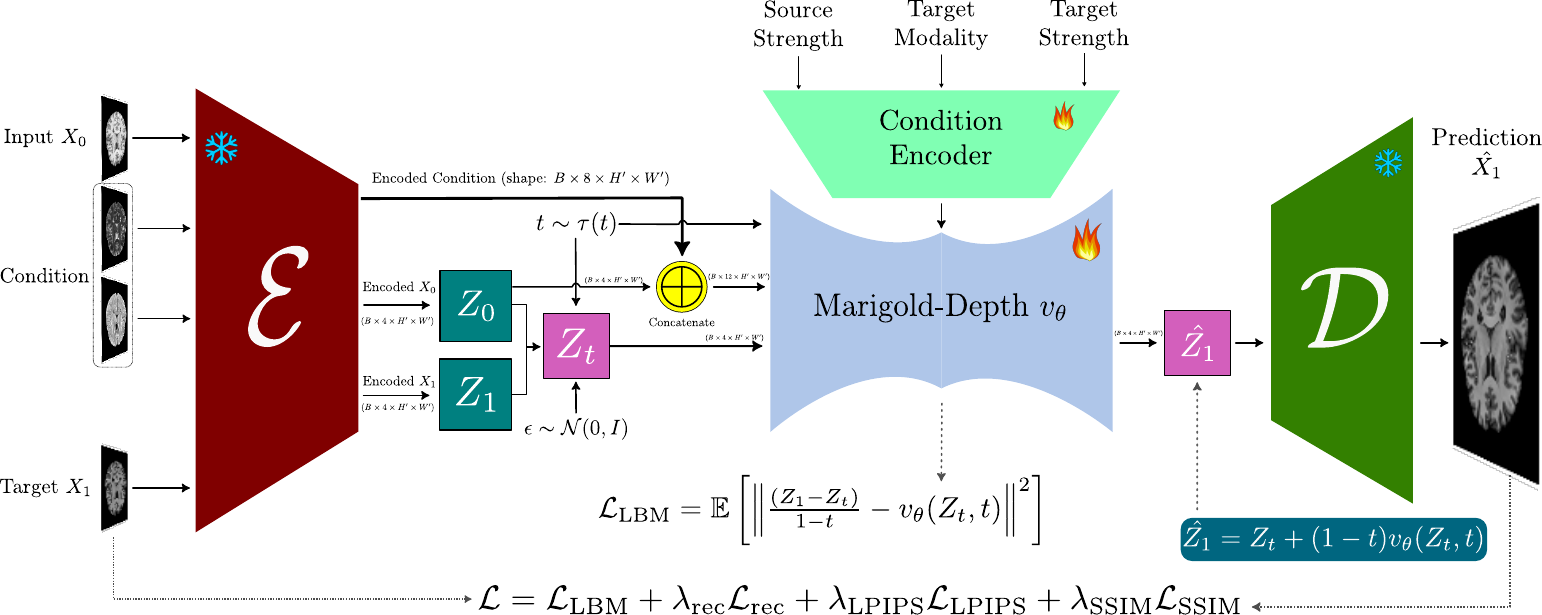}
    \caption{Training conditional latent bridge matching model for MRI synthesis.}
    \label{fig:marigold-lbm}
\end{figure}

\paragraph{Bridge Model.} 
The drift function $v_\theta$ is initialised from the Marigold-Depth \cite{kemarigold2024} model, specifically the \texttt{prs-eth/marigold-depth-v1-1} checkpoint; Marigold is itself based on the Stable Diffusion (SD) v2 \cite{rombach2022highresolutionimagesynthesislatent} UNet \cite{ronneberger2015unetconvolutionalnetworksbiomedical} and adapted to perform image-to-image translation for various image analysis tasks \cite{kemarigold2024}. We extend Marigold's input convolutional layer to accept a $16$-channel latent, corresponding to four latent groups of four channels each: the input image, the target image, and the two auxiliary modality condition slices, shown in \Cref{fig:marigold-lbm}. We initialise the expanded convolution and replicate the weights of the first four input channels threefold and rescale by a factor of $1/3$; the weights of the remaining four channels are copied without modification. This initialisation preserves the network's response at the start of training and is consistent with the initialisation used in Marigold \cite{kemarigold2024}. $v_{\theta}$ is optimised end-to-end using the objective
\begin{equation}
    \mathcal{L} = \mathcal{L}_{\text{LBM}} + \lambda_{\text{rec}} \mathcal{L}_{\text{rec}} + \lambda_{\text{LPIPS}} \mathcal{L}_{\text{LPIPS}} + \lambda_{\text{SSIM}} \mathcal{L}_{\text{SSIM}}
    \label{eqn:full_bridge_loss}
\end{equation}
where $\mathcal{L}_{\text{LBM}}$ is the LBM loss defined in \cref{eqn:lbm_loss}, $\mathcal{L}_{\text{rec}}$ is the reconstruction loss, $\mathcal{L}_{\text{LPIPS}}$ is the LPIPS \cite{zhang2018perceptual} loss, and $\mathcal{L}_{\text{SSIM}}$ is the structural similarity index measure (SSIM) \cite{wangbovik2004} loss (defined as $1 - \text{SSIM}$). 

\paragraph{Condition Encoder.}
The Stable Diffusion UNet backbone of Marigold is a text-to-image model and hence utilises additional conditioning inputs to guide the image generation process using cross-attention. We leverage this setup  and design a lightweight condition encoder to encode the source strength (i.e. the field strength of the input $X_0$), the target modality and the target strength to guide the cross-strength translation. The source and target strengths, $s^{\text{src}}, s^{\text{tgt}}$, are first normalised to $[0,1]$ by the maximum field strength in the dataset and mapped to fixed sinusoidal encodings 
\[ \gamma(s) = \left[\sin (s \cdot \omega_k); \cos (s \cdot \omega_k)\right]_{k=0}^{K-1} \in \mathbb{R}^{2K}, \omega_k = \frac{2 \pi k}{K-1} \]
with $K = 8$. The resulting vectors $\gamma(s^{\text{src}})$ and $\gamma(s^{\text{tgt}})$ are then independently projected to cross-attention space ($\mathbb{R}^{1024}$ in the case of SD 2.1) through a learnt linear map. The target modality is projected to the same cross-attention space using a learnt embedding table. The three embeddings are stacked to form the condition embedding, which is appended to the cross-attention context of $v_{\theta}$. 

\subsection{Training}
\paragraph{VAE Pretraining.}
We first train a variational autoencoder on the large unpaired dataset to encode and decode samples to latent space, following the paradigm of latent diffusion \cite{rombach2022highresolutionimagesynthesislatent}. The loss combines the reconstruction, KL, and perceptual terms from \cref{eqn:full_vae_loss}: $\lambda_{\text{rec}}=0.5$ with $\mathcal{L}_{\text{rec}}=L^1$, $\lambda_{\text{KL}}=1\times10^{-6}$, and $\lambda_{\text{LPIPS}}=0.1$, where $\mathcal{L}_{\text{LPIPS}}$ uses the VGG network \cite{simonyan2015a} applied patch-wise with patch size $64$ and stride $64$. We train for $25{,}000$ optimiser steps with AdamW \cite{loshchilov2019decoupledweightdecayregularization}, batch size $4$, and $4$ gradient accumulation steps, using $2{,}500$ linear warm-up steps to a peak learning rate of $3\times10^{-5}$. Weights are tracked with an exponential moving average (EMA) ($\beta=0.9997$). We train on $3$ NVIDIA A10 GPUs for a total of $28$ hours.

\paragraph{Drift Model.}
We train the drift model and condition encoder on the smaller paired dataset using the bridge matching loss from \cref{eqn:full_bridge_loss} with $\sigma=0.001$ and the weighted time schedule $\tau(t)=0.9\delta_{t=0}+0.025\delta_{t=0.25}+0.05\delta_{t=0.5}+0.025\delta_{t=0.75}$. The loss weights nRMSE reconstruction ($\lambda_{\text{rec}}=0.1$), SSIM ($\lambda_{\text{SSIM}}=0.1$), and AlexNet \cite{krizhevsky2012alexnet} LPIPS \cite{zhang2018perceptual} ($\lambda_{\text{LPIPS}}=0.2$). The pipeline trains end-to-end with AdamW \cite{loshchilov2019decoupledweightdecayregularization}, batch size $2$ and $16$ gradient accumulation steps, over $500$ linear warm-up steps followed by 10{,}000 training steps. The drift model $v_\theta$ uses a learning rate of $3\times10^{-5}$; the condition encoder uses $1.5\times10^{-4}$. We apply ZeRO-1 \cite{deepspeed2020}, mixed precision, and an EMA of the weights ($\beta=0.9995$). Our model trains on $3$ NVIDIA A10 GPUs for $40$ hours.

\subsection{Inference}
At inference we are given three source contrasts (T1, T2, and T2FLAIR) of a subject, all imaged at field strength $s^\text{src}$,  and we wish to generate a volume of the same subject at target modality $m^\text{tgt}$ and target strength $s^\text{tgt}$. We generate each slice independently along the axial axis by first centre-cropping the three source contrasts to $384 \times 384$ and encoding each to latent space with the VAE encoder, producing three $4$-channel latents that are concatenated to form the $12$-channel source tensor $Z_\text{src}$. $v_\theta$ operates on a $16$-channel input formed by concatenating this source tensor with the bridge state $Z_t$ from \cref{eqn:br_bridge}; however, since $Z_1$ is unavailable at inference (we wish to predict this target latent), we simply initialise the bridge state at $Z_0$, i.e. $Z_t := Z_0$. The triplet ($s^\text{src}$, $m^\text{tgt}$, $s^\text{tgt}$) is mapped by the condition encoder to a condition embedding, which together with $Z_\text{src}$ is fixed throughout integration. The bridge integration proceeds as per \cref{sec:inference} on [$Z_\text{src}; Z_t$]; after $N$ steps, the predicted target latent $\hat{Z}_1$ is mapped back to pixel space using the VAE decoder, and finally zero-padded back to the original resolution.

\section{Results}
\begin{figure}[t!]
    \centering
    \includegraphics[width=1\linewidth]{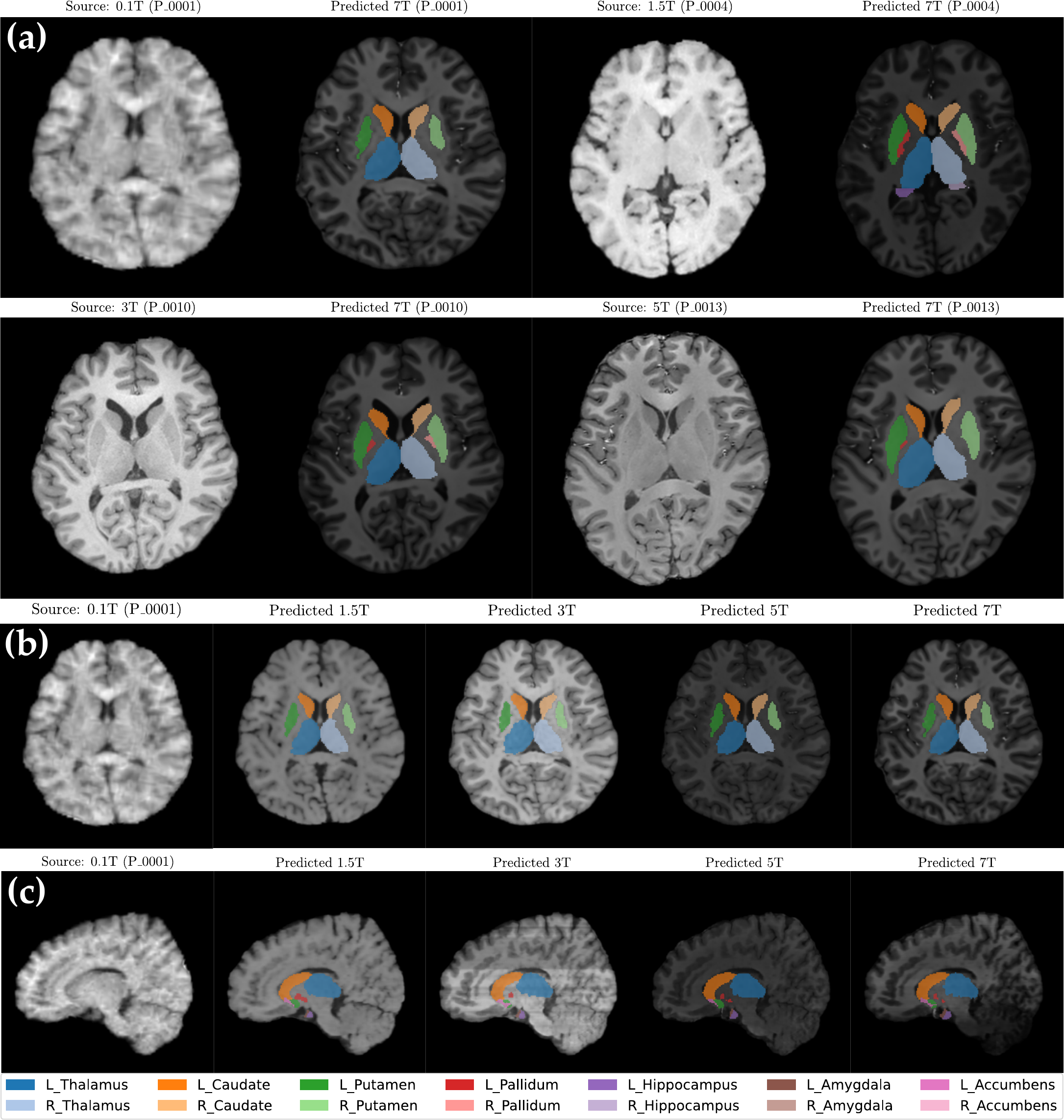}
    \caption{Results for Task 1 and Task 2 using our single unified model. \textbf{(a)} T1W model outputs and deep grey matter (DGM) segmentations for ultra-high-field synthesis from low-field 2D slices (Task 1, $z = 160$). \textbf{(b)} Axial T1W model outputs and DGM segmentations for higher-field image synthesis from ultra-low-field slices (Task 2, $z = 160$). \textbf{(c)} Sagittal T1W outputs and DGM segmentations for higher-field image synthesis from ultra-low-field slices (Task 2, $z = 160$).} 
     \label{fig:3}
    \begingroup
    \renewcommand{\thefigure}{\arabic{figure}\alph{subfigure}}
    \setcounter{subfigure}{1}  
    \label{fig:task1-figure}
    \setcounter{subfigure}{2}  
    \label{fig:task2-figure}
    \setcounter{subfigure}{3}
    \label{fig:task2-sagittal}
    \endgroup
\end{figure}

We present our validation results for all three tasks of the MRIxFields2026 challenge in \Cref{table:results_table}. Our unified model is competitive across all metrics on all three tasks on the validation leaderboard. We show outputs of our model and deep grey matter (DGM) segmentations in \Cref{fig:3}. \Cref{fig:task1-figure} showcases ultra-high-field (7T) synthesis from low-field slices for four subjects on the validation dataset, overlaid with their corresponding SynthSeg segmentations \cite{BILLOT2023102789}; \Cref{fig:task2-figure} showcases higher-field synthesis from a 0.1T volume for a single input subject. Our synthesis enhances detail and contrast in each of the slices with increasing field strength and preserves the size and location of DGM structures across all higher-field slices. Lastly, \Cref{fig:task2-sagittal} shows that since our method generates each 2D axial slice of a volume independently, inter-slice intensity consistency is reduced, shown as banding artefacts in sagittal sections. It is possible to reduce thse artefacts by generating volumes along the three spatial orientations and averaging; this improves every metric (e.g. lowers nRMSE from 0.289 to 0.266 on Task 1, 0.202 to 0.188 on Task 2, 0.230 to 0.219 on Task 3) except LPIPS (0.072 to 0.077 on Task 1, 0.092 to 0.103 on Task 2, 0.081 to 0.083 on Task 3) at the cost of triple the inference time per volume.

\setlength{\intextsep}{0pt}
\begin{wraptable}[8]{r}{7.3cm}
\centering
\caption{Validation metrics.}
\begin{tabular}{cccccc}
\hline
\cellcolor[HTML]{EFEFEF}                                & \cellcolor[HTML]{EFEFEF}                                & \cellcolor[HTML]{EFEFEF}                                 & \cellcolor[HTML]{EFEFEF}                                & \cellcolor[HTML]{EFEFEF}                                & \cellcolor[HTML]{EFEFEF}                                  \\
\multirow{-2}{*}{\cellcolor[HTML]{EFEFEF}\textbf{Task}} & \multirow{-2}{*}{\cellcolor[HTML]{EFEFEF}\textbf{SSIM}} & \multirow{-2}{*}{\cellcolor[HTML]{EFEFEF}\textbf{LPIPS}} & \multirow{-2}{*}{\cellcolor[HTML]{EFEFEF}\textbf{nRMSE}} & \multirow{-2}{*}{\cellcolor[HTML]{EFEFEF}\textbf{Dice}} & \multirow{-2}{*}{\cellcolor[HTML]{EFEFEF}\textbf{Volume}} \\ \hline
Task 1                                                  & 0.914                                                   & 0.072                                                    & 0.289                                                   & 0.898                                                   & 0.880                                                     \\ \hline
Task 2                                                  & 0.885                                                   & 0.092                                                    & 0.202                                                   & 0.857                                                   & 0.839                                                     \\ \hline
Task 3                                                  & 0.902                                                   & 0.081                                                    & 0.230                                                   & -                                                       & -                                                         \\ \hline
\end{tabular}
\label{table:results_table}
\end{wraptable}

\subsection{Ablations}
\begin{figure}[t]
    \centering
    \begin{minipage}{0.49\linewidth}
    \includegraphics[width=\linewidth]{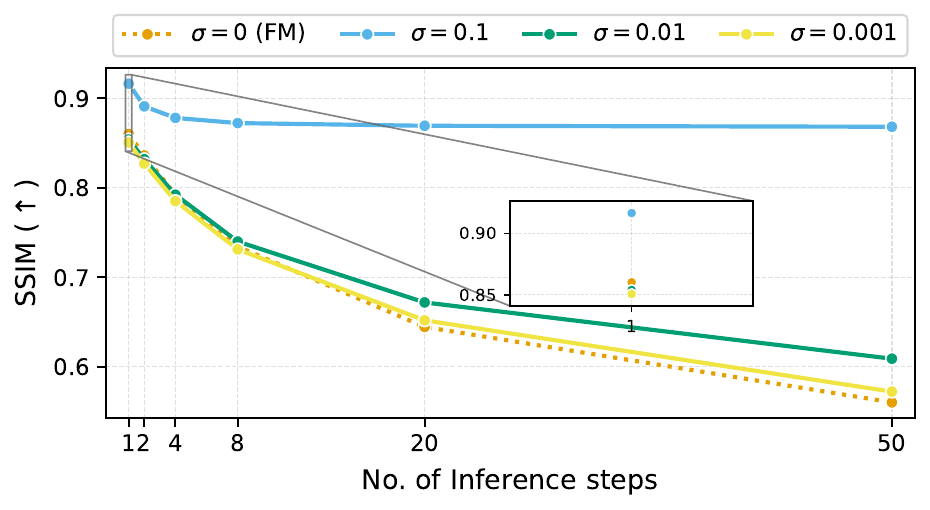}
    \subcaption{Uniform distribution $\tau_\mathcal{U}(t)$}
    \label{fig:abl-uniform}
    \end{minipage}
    \begin{minipage}{0.49\linewidth}
    \includegraphics[width=\linewidth]{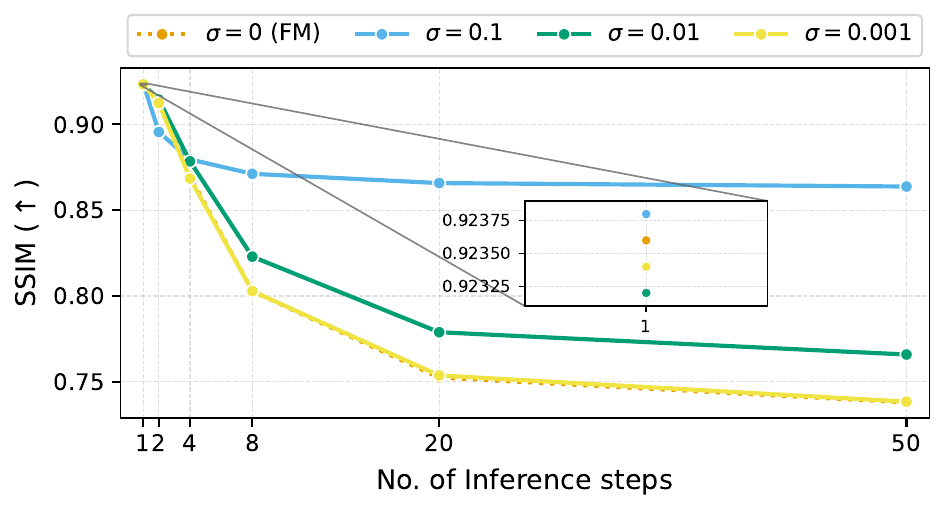}
    \subcaption{Discrete Uniform $\tau_{\text{D}\mathcal{U}}(t)$}
    \label{fig:abl-discrete-uniform}
    \end{minipage}\\

    \begin{minipage}{0.49\linewidth}
    \includegraphics[width=\linewidth]{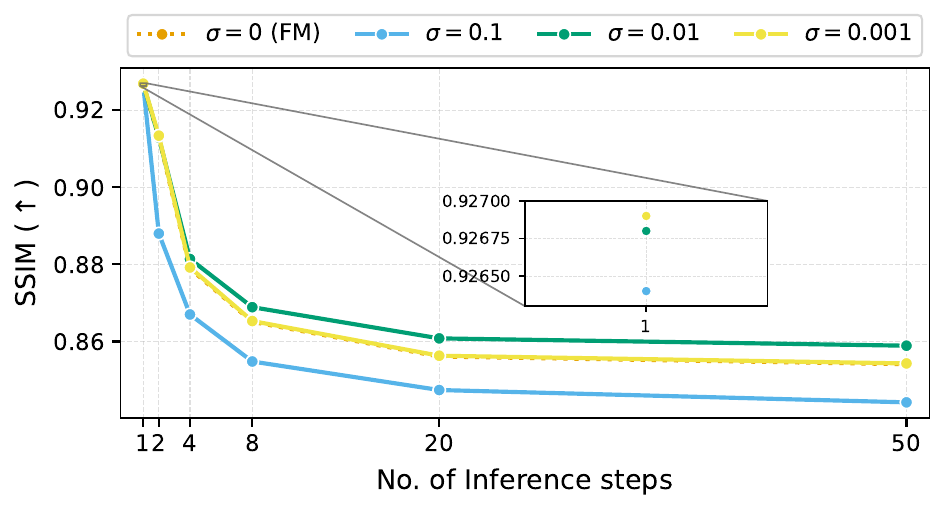}
    \subcaption{Discrete Weighted $\tau_{\text{DW}}$(t)}
    \label{fig:abl-discrete-weighted}
    \end{minipage}
    \begin{minipage}{0.49\linewidth}
    \includegraphics[width=\linewidth]{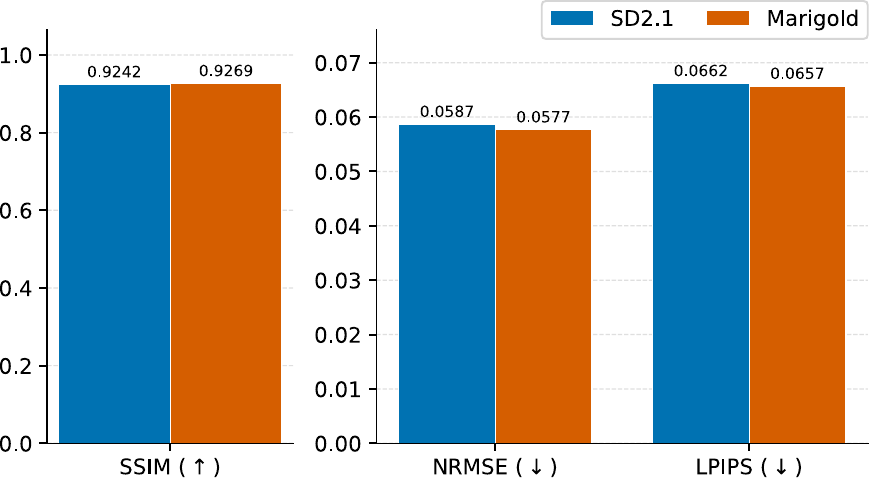}
    \subcaption{Marigold vs SD2.1}
    \label{fig:abl-marigold-sd21}
    \end{minipage}
    \caption{Ablation across $\sigma$, noise schedules $\tau(t)$, and number of inference steps.}
    \label{fig:ablations}
\end{figure}
For our ablations, we train all models for $5,000$ steps with $250$ linear warm-up steps to a peak learning rate of $3 \times 10^{-5}$ and track the weights with an EMA ($\beta = 0.9995$) using the Marigold UNet. We use our pretrained VAE for all experiments and only focus on the drift model $v_\theta$; without loss of generality, we train on 2 out of 3 volunteers in the paired dataset and validate on the held-out volunteer. Our best model uses the Marigold UNet with $\sigma = 10^{-3}$ and the discrete-weighted schedule, achieving SSIM = $0.9269$ in 1 inference step.

\paragraph{Impact of timestep distribution $\tau(t)$.} We test three choices of the timestep distribution $\tau(t)$: a uniform distribution $\tau_{\mathcal{U}}(t) = \mathcal{U}([0,1))$, a discrete-weighted distribution $\tau_{\text{DW}}(t)=0.9\delta_{t=0}+0.025\delta_{t=0.25}+0.05\delta_{t=0.5}+0.025\delta_{t=0.75}$ (from the original LBM paper \cite{chadebecLBMLatentBridge2025}), and lastly a discrete-uniform distribution $\tau_{\text{D}\mathcal{U}}(t) = \mathcal{U}(\{0, 0.25, 0.5, 0.75\})$. The results are shown in \Cref{fig:abl-uniform,fig:abl-discrete-uniform,fig:abl-discrete-weighted}.

\paragraph{Impact of bridge noise $\sigma$.} We ablate values for the bridge noise $\sigma$ in \cref{eqn:latent_interpol}: we test $\sigma = 0, 0.1, 0.01, 0.001$. $\sigma=0$ is the flow-matching (FM) instantiation. The results are shown in \Cref{fig:abl-uniform,fig:abl-discrete-uniform,fig:abl-discrete-weighted}.

\paragraph{Impact of model $v_\theta$.} We test the impact of two different choices of $v_\theta$: the original SD $2.1$ UNet from the \texttt{sd-research/stable-diffusion-2-1-base} checkpoint, as well as the Marigold-Depth model (\Cref{fig:abl-marigold-sd21}). The SD $2.1$ UNet has its \texttt{conv\_in} weights replicated $4$ times and rescaled by $1/4$ to accept a $16$-channel latent.

\section{Conclusion}
We present a single, unified model addressing all three tasks of the MRIxFields2026 challenge, based on conditional latent bridge matching. We can improve our method by generating volumes along three spatial orientations and averaging, producing improved distortion metrics at the expense of perceptual metrics, and at the cost of higher inference time.

\begin{credits}
\subsubsection{\ackname}
The authors acknowledge the use of the Batch Compute System in the Department of Computer Science at the University of Warwick, and associated support services, in the completion of this work. Calculations were performed using the Sulis Tier 2 HPC platform hosted by the Scientific Computing Research Technology Platform at the University of Warwick. Sulis is funded by EPSRC Grant EP/T022108/1 and the HPC Midlands+ consortium. Computing facilities were provided by the Scientific Computing Research Technology Platform of the University of Warwick. S.S. is funded by the Department of Computer Science, University of Warwick. T.B. is supported through EPSRC/NSF grant EP/X026663/1. 
\end{credits}

\bibliographystyle{splncs04}
\bibliography{refs}

\end{document}